\documentclass[11pt,a4paper]{article}
\usepackage[hyperref]{acl2020}
\usepackage{times}
\usepackage{latexsym}
\usepackage{multirow}
\usepackage{graphicx}
\usepackage{amsmath}
\usepackage{cleveref}
\usepackage{amsfonts}
\usepackage{amsthm}

\theoremstyle{definition}

\usepackage{microtype}

\aclfinalcopy %
\def\aclpaperid{41} %

\makeatletter
\newcommand*\bigcdot{\mathpalette\bigcdot@{.5}}
\newcommand*\bigcdot@[2]{\mathbin{\vcenter{\hbox{\scalebox{#2}{$\m@th#1\bullet$}}}}}
\makeatother

\newcommand{\norm}[1]{\left\lVert#1\right\rVert}

\crefformat{section}{\S#2#1#3} %
\crefformat{subsection}{\S#2#1#3}
\crefformat{subsubsection}{\S#2#1#3}

\title{Staying True to Your Word: (How) Can Attention Become Explanation?}

\author{Martin Tutek \and Jan \v{S}najder\\
Text Analysis and Knowledge Engineering Lab\\
Faculty of Electrical Engineering and Computing, University of Zagreb\\
Unska 3, 10000 Zagreb, Croatia \\
\tt \{martin.tutek,jan.snajder\}@fer.hr
}
\date{}

\begin{document}
\maketitle
\begin{abstract}
The attention mechanism has quickly become ubiquitous in NLP.
In addition to improving performance of models, attention has been widely used as a glimpse into the inner workings of NLP models.
The latter aspect has in the recent years become a common topic of discussion, most notably in work of \citealp{jain2019attention, wiegreffe2019attention}.
With the shortcomings of using attention weights as a tool of transparency revealed, the attention mechanism has been stuck in a limbo without concrete proof when and whether it can be used as an explanation.
In this paper, we provide an explanation as to why attention has seen rightful critique when used with recurrent networks in sequence classification tasks.
We propose a remedy to these issues in the form of a word level objective and our findings give credibility for attention to provide faithful interpretations of recurrent models.

\end{abstract}

\section{Introduction} %
\label{sec:introduction}

Not long since its introduction, the attention mechanism \cite{bahdanau2014neural} has become a staple of many NLP models.
Apart from enhancing prediction performance of models and starting the trend of fully attentional networks
\cite{vaswani2017attention}, attention weights have been widely used as a method for interpreting decisions of neural models.

Recently, the validity of interpreting the decision making process of a model through its attention weights came under question.
\newcite{jain2019attention} introduced a set of experiments on English language sequence classification tasks which demonstrated that attention weights do not correlate with feature importance measures, and that attention weights generated by a trained model can be substituted and modified without detriment to model performance.
While it is natural to assume that multiple plausible explanations for a model's decision can coexist, the authors show the existence of attention distributions that assign most of their mass to words seemingly irrelevant to the task, while still not affecting neither the decision nor the confidence of the model.
In the follow-up work, \newcite{wiegreffe2019attention} find that, while such \textit{adversarial} attention distributions do exist, they are seldom converged to in the training process, even when one introduces a training signal with the sole purpose of guiding the model to such distributions.

In this paper, we aim to tackle the difficult question of the relationship between attention and explanation from a different angle -- is there any modification we can make to the existing models so that attention could be reliably used as a tool of model transparency? 
For the sake of consistency, we follow previous work \cite{jain2019attention, wiegreffe2019attention} and limit our scope to single-sequence binary classification tasks, where we consider models from the \textit{RNN + self-attention} family.
Concretely, we analyse single-layer bidirectional LSTM-s \cite{hochreiter1997long} equipped with the additive \cite{bahdanau2014neural} and dot-product \cite{vaswani2017attention} self-attention mechanisms.

Inspired by the recent results \cite{voita2019bottom}, which show that optimizing the masked language modelling (MLM) \cite{devlin2019bert} objective results in high mutual information between the input and output layers of models, we ask ourselves whether such a trait is beneficial for interpretability.
The task of sequence classification in no way incentivizes a model to retain information from the input, and the model is likely to filter out information irrelevant to the task.\footnote{The LSTM cell even has an inductive bias towards forgetting information, as we cannot expect the cell gates to always be saturated on the positive side.} 
We believe this lack of enforced information retention causes a discrepancy between the input and hidden vectors, which results in reduced model interpretability. 
To enforce information retention, we propose a number of techniques to keep the hidden representations closer to their input representations, improving the faithfulness of interpreting models through inspecting their attention weights.

The contributions of this paper are as follows: we (1) investigate whether the
lack of a word-level objective causes attention not to be a faithful
interpretation, (2) propose various regularization methods in order to improve interpretability through inspecting attention weights,
and (3) quantitatively and qualitatively evaluate whether and how these methods help model interpretability.

The rest of the paper is organized as follows. Firstly (\cref{sec:related_work}), we position ourselves within current work and discuss the use of attention as interpretation in NLP,.
We then (\cref{sec:methodology}) present our experimental setup, introduce various regularization methods, and briefly describe the experiments we use to evaluate our regularized models.
In \cref{sec:experiments}, we offer a quantitative evaluation of the effect of regularizes on the trained models across a number of datasets.
We then (\cref{sec:effect_of_masked_language_modelling}) qualitatively and quantitatively inspect the effect of regularization on a trained model, identifying what we believe to be the cause of negative results reported in previous work.
Finally (\cref{sec:conclusion}), we summarize our findings and propose possible lines of future work.

\section{Attention and Interpretability in NLP} %
\label{sec:related_work}

\textbf{Preliminaries}:
Let the input sequence of word embeddings be denoted as $\{w_t\}_{t=1}^T$, where $T$ is the length of the sequence.
The sequence of hidden states produced by the encoder is then $\{h_t\}_{t=1}^T$, where each $h_t = \textrm{rnn}(x_t, h_{(t-1)})$. The RNN used is a bidirectional LSTM.
When discussing a hidden state $h_t$, we refer only to $x_t$ as its \textit{input} for convenience.
The attention mechanism produces a probability distribution over the hidden states, the elements of which we denote $\{\alpha_t\}_{t=1}^T$, and refer to as \textit{attention weights}.

\subsection{Attention as Interpretation} %
\label{sub:attention_as_interpretation}

When interpreting models through the attention mechanism, we assume that the attention weight on the $t$-th word, $\alpha_t$, is a \textit{faithful} measure of importance of the input word $x_t$ for the classifier decision.
This assumption allows us to \textit{interpret} the decision of the classifier by retrieving the highest attention weights assigned by the model, and then
identifying the input words in these timesteps. 
Thus, in the terminology of \newcite{doshi2017towards}, our \textit{cognitive chunk} (a basic unit of explanation) is a single word.
However, we are using a BiLSTM as an encoder, and every hidden state is contextualized by virtue of observing the entire input sequence, so the attention weights actually pertain to the input word \text{in context}.
A faithful measure of importance should by definition accurately represent the true reasoning behind the final decision of the model.\footnote{For an excellent discussion on interpretation faithfulness, see Alon Jacovi's post on \url{https://tinyurl.com/y92rskfr}}
So, if attention weights are a faithful measure of importance of word inputs, they will assign large weights to words relevant for the classifier decision.

To define faithfulness more clearly, we can assume the existence of an oracle method which can partition each input sequence of words\footnote{The instance-level definition is important here, as the same word can bear different meanings in different contexts.} into \textit{decision-relevant} and \textit{decision-irrelevant}
words, where relevance is defined by the judgment of a human reading the text with respect to a task.
By this definition, a faithful attention distribution would consistently attribute all or at least most of its probability mass to the decision-relevant words, making it a \textit{plausible} explanation for humans. 
In contrast, a \textit{counterfactual} attention distribution \cite{jain2019attention} attributes most (or a significant amount) of its probability mass to task-irrelevant words.
Obviously, infinitely many plausible and counterfactual explanations exist for a given input instance -- merely by redistributing the original attention mass within the same set of words we can obtain infinitely many \textit{alternative} interpretations that are still either plausible or counterfactual.

\newcite{jain2019attention} and \newcite{vashishth2019attention} demonstrate that, if we permute or substitute the weights of a learned attention distribution, our model can still retain high (and in some cases, unchanged) classification performance and prediction confidence.
Even more worryingly, some of the modified attention distributions assign high attention weights to task-irrelevant words while not affecting the instance classification.
The existence of such counterfactual attention distributions raises doubts whether inspecting attention weights can be used as a faithful interpretation of the model's decision making process at all.

\newcite{wiegreffe2019attention} provide two counter-arguments -- (1) \textit{Existence does not entail exclusivity}, suggesting that, just because our model has converged to an attention distribution (a \textit{base} attention distribution), that distribution is not necessarily unique, and alternative attention distributions can still be faithful; (2) while models which produce counterfactual distributions do exist and can be found by post-hoc modifications, these distributions are difficult to converge to naturally through the optimization process of a neural network. This is demonstrated by the authors in experiments where they specifically optimize for a distribution significantly different from the base one.

In contrast, \newcite{rudin2019stop} states that even if a small fraction of explanations produced by the model is counterfactual, one cannot trust other explanations produced by the same model. \newcite{lipton2016mythos} is more forgiving, and allows that models can still be trusted if they make mistakes, provided humans would also make mistakes on the same instances.
The work of \newcite{pruthi2019learning} emphasizes the threat of interpreting models through attention weights, as they show a regularization term can be introduced to guide the attention weights away from focusing on subsets of words while retaining model accuracy, implying that models which exploit bias in data can be trained to hide the true reasoning behind their decisions.

Among other work, \newcite{serrano2019attention} apply an array of tests to analyse whether attention weights correlate with impact on model prediction, concluding again that attention is not a fail-safe (faithful) indicator of importance. The experiments of \newcite{vashishth2019attention} show that for single-sequence classification, learned attention distributions can be replaced without affecting performance -- indicating that attention might not be all we need, after all.

\section{Experimental Setup} %
\label{sec:methodology}

The \textbf{base} model used in \cite{jain2019attention, wiegreffe2019attention} is a single-layer bidirectional LSTM augmented with either a dot-product or an additive attention mechanism, the output of which is then fed into a linear classifier (decoder).
We use the same base model as a baseline throughout our experiments.

\subsection{Regularizing Models} %
\label{sub:regularizing_models}

As mentioned before, we suspect that the lack of a word-level objective weakens the relationship between $h_t$ and $x_t$, and, consequently, the faithfulness of interpreting attention weights $\alpha_t$ as an explanation of the decision making process of the model diminishes.
We will now present a number of methods constructed with the goal of improving information retention between the inputs and hidden states.

Our self-attention augmented LSTM encoder with inputs $x_t$ is defined as:
\begin{align}
\label{eq:base_lstm}
\begin{gathered}
	e_t = \mathrm{emb}(x_t) \\
 h_t = \mathrm{rnn}(e_t)
\end{gathered}
&&
\begin{gathered}
\alpha_t = \mathrm{attn}(h_t) \\
 s = \sum \alpha_i h_i
\end{gathered}
\end{align}
\noindent where $\mathrm{attn}$ is either the dot-product or additive attention mechanism.
The sequence representation $s$ is then fed into a linear decoder.

The simplest way to retain information from input is to include it explicitly in the hidden representations.
This can be done by concatenating the embeddings to the hidden representation:
\begin{equation}
	h^{\mathit{cat}}_t = [\mathrm{rnn}(e_t);e_t]
\end{equation}
\noindent where $[\cdot;\cdot]$ is the concatenation operator.
Another method is to incorporate a residual connection:

\begin{equation}
	h^{\mathit{res}}_t = e_t + \mathrm{rnn}(e_t)
\end{equation}

We use these two methods as our regularized baselines (\textbf{concat}, \textbf{residual}), along with the unreguralized \textbf{base} model.

Our next proposed method is to add a regularization term constraining the L2 norm of the difference between a word embedding and its corresponding hidden representation.
As we suspect that the base model discards a lot of word information it deems task-irrelevant, we wish to penalize it for doing so where this information filtering is not crucial.

\begin{equation}
	\mathcal{L}_{\mathit{tying}} = \frac{\delta}{T} \sum_i^T \norm{h_t - e_t}_2^2
\end{equation}

\noindent where $\delta$ is the regularization scale hyperparameter, and we minimize the average across all tokens in the batch.
We consider values $[1,10, 20, 30]$ for $\delta$ and perform ablation for these values.
Further on, we only report results of the model with the best-performing results due to space limitations.
We further refer to this method as \textbf{tying}.

The last model we propose is inspired by results in \cite{voita2019bottom}, where we introduce the masked language modelling objective \cite{devlin2019bert}, in which input tokens from a sequence are masked at random.\footnote{To be precise, either masked, replaced by a random word, or left unchanged. We direct the reader to \cite{devlin2019bert} for a detailed explanation of the MLM task.}  
The task of the model is then to correctly predict the masked tokens based on contextual cues from the unmasked tokens in the sequence.

In addition to the standard model in \eqref{eq:base_lstm}, the \textbf{MLM} model also performs the following:

\begin{align}
	\hat{x_t} &= \mathrm{mask}(x_t) \nonumber \\
	\hat{e}_t &= \mathrm{emb}(\hat{x}_t) \\
	\hat{h}_t &= \mathrm{rnn}(\hat{e}_t) \nonumber
\end{align}
The hidden states $\hat{h}_t$ for the corresponding masked tokens are then fed into a linear decoder which predicts the masked word.
The encoder and embedding matrix are shared between the MLM and classification tasks.

The MLM linear decoder also introduces no new parameters as we tie the weights \cite{inan2016tying} of the MLM decoder and the input embedding matrix and keep them frozen during training.
Both of these choices are motivated by the fact that the model might converge to a solution which does not require retention of information from inputs.
In order to apply weight tying, we have to ensure that the dimension of the BiLSTM hidden state equal to the input embedding, and therefore we increase the LSTM hidden state size to $150$, compared to $128$ in \cite{jain2019attention,
wiegreffe2019attention}.
We also use the new hidden state size for all experiments with the base model.

The MLM setup introduces two hyperparameters: $p_{\mathit{mlm}}$, denoting the
probability of masking a token in a sequence, and $\eta$, denoting the
weight of the MLM loss. We keep $p_{\mathit{mlm}}$ fixed at $0.15$ throughout the
experiments, as in \cite{devlin2019bert}, and adjust $\eta$ with respect to the average sequence length in various datasets so that the MLM loss would not dominate
the optimization process.\footnote{As due to keeping $p_{\mathit{mlm}}$ fixed, the longer the sequence is, the more masked predictions we are expected to make.}

\subsection{Post-hoc Modification of Attention Distributions} %
\label{sub:attention_distributions}

As suggested by \newcite{jain2019attention}, robustness of classifier confidence with respect to attention weight modifications is not a desirable property of interpretable models.
Ideally, if a model produces the same decision for an alternative set of attention weights, we would like to be sure that the alternative explanation is faithful.
This is not the case in practice as \newcite{jain2019attention} and \newcite{vashishth2019attention} show that a trained network is surprisingly robust to changes to the attention weights and produces nearly unchanged classification scores even for adversarial distributions.
So, while attention is an integral part of training the network, the weights it produces do not greatly affect the classifier decision once trained.

While we agree with the observation of \newcite{wiegreffe2019attention} that robustness of model decisions with respect to attention weights is not necessarily bad as the model is unlikely to naturally converge to such a solution, we believe that fragility of model decisions is an argument \textbf{in favor} of interpretability as it indicates that the number of explanations plausible to the model has been reduced, and we perform experiments with that in mind.

\subsection{Training an Adversary} %
\label{sub:training_adversary}

In the experiment introduced by \newcite{jain2019attention}, for a trained model we attempt to find an adversarial attention distribution which maximizes the Jensen-Shannon divergence (JSD) from the base distribution produced by the trained model, while at the same time minimizing the total variation distance (TVD) from the confidence of the predictions of the base model.
The authors demonstrate that it is possible to find an attention distribution that obtains a high JSD while still producing the same prediction confidence consistently across multiple tasks.

As these adversarial distributions were found in an artificial setting, \newcite{wiegreffe2019attention} explore a more realistic scenario and construct an optimization task where, given a fixed (original) model, they train an adversary to minimize TVD from per-instance prediction confidences, while maximizing JSD between per-instance attention distributions of the original model and the adversary.
The optimization objective for our adversarial model $a$ given a base model $b$ is defined as follows:
\begin{equation}
  \mathcal{L} = \mathrm{TVD}(\hat{y}_a, \hat{y}_b) - \lambda \mathrm{JSD}(\alpha_a, \alpha_b)
\end{equation}
This training setup introduces another hyperparameter $\lambda$, which weighs the JSD component of the optimization objective.
TVD and JSD are defined as follows:
\begin{align}
	\mathrm{TVD}(\hat{y}_a, \hat{y}_b) &= \frac{1}{2} \sum_{i=1}^{\left\vert\mathcal{Y}\right\vert} \left\vert y_{ai} - y_{bi} \right\vert\\
  \mathrm{JSD}(\alpha_a, \alpha_b) &= \frac{1}{2} (\mathrm{KL}[\alpha_a || \bar{\alpha}]  + \mathrm{KL}[\alpha_b || \bar{\alpha}])
\end{align}
where $\bar{\alpha} = \frac{\alpha_a + \alpha_b}{2}$.

Initially, we were enthusiastic about this setup and conducted the same experiments with our model variants, but drawing any conclusions from the analysis proved to be hard.
Firstly, by optimizing for TVD from a trained model instead of on the raw labels, we bias our new model to make the exact same mistakes as the trained model.
We believe this severely limits the search space of the adversarial model, as repeating the same mistakes will also bias the model towards exploiting similar patterns in data and, consequently, a similar attention distribution.
Secondly, without knowing what the plausible explanations are for the dataset, it is impossible to determine whether a high JSD is a symptom of the model finding an alternative or adversarial explanation.
Thus, we do not attempt to draw many conclusions from this experiment, but we reproduce it for completeness with previous work.

\subsection{Mutual Information} %
\label{sub:mutual_information}
To quantitatively evaluate whether the regularization has strengthened the relationship between the hidden states and input representations of our model, we look into a recent method of \newcite{voita2019bottom} inspired by the ``Information Bottleneck'' (IB) theory \cite{tishby1999information}, where the authors measure an estimate of mutual information (MI).
Originally applied to transformers \cite{devlin2018bert}, this method is straightforward to adapt to the bidirectional LSTM.

Similarly to our point of view, the IB theory states that neural networks, in general, aim to extract a compressed representation of input in which information relevant for the output is retained while irrelevant is discarded.
Mutual information is used as a method of measuring how much information is lost between the input and hidden representation of a certain network.
\newcite{voita2019bottom} show transformer networks discard progressively more information in deeper layers.
This phenomenon is different for the case of MLM in transformers, where MI is higher in the uppermost layers, likely due to the task of reconstructing corresponding input tokens.

The strength of the relationship between $e_t$ and $h_t$ can be quantified by estimating MI.
As MI is intractable to compute in the continuous form, we first discretize the vector representations and estimate MI in the discrete form.
Following \newcite{voita2019bottom} and \newcite{sajjadi2018assessing}, we perform this discretization by clustering the embedding and hidden representations to a large number of clusters and using the obtained cluster labels in place of the continuous vectors to estimate MI.

Concretely, we select a subset of 1000 words from the vocabulary and gather at most 1M representations of these tokens at input and hidden level.
We then cluster the obtained representations into $k=1000$ clusters with mini-batch $k-$means with batch size of 100.
We obtain the vocabulary sample in two ways: as the top 1k most frequent words (\textbf{MF}), as in \cite{voita2019bottom}, but also as a random sample (\textbf{RS}) of from the scaled unigram distribution.\footnote{The sample is drawn from the unigram distribution raised to the power of $\frac{3}{4}$.}

\subsection{Datasets} %
\label{sub:datasets}

We experiment on the following English language datasets for binary classification tasks, which were either originally built for this task or were adapted for it by \newcite{jain2019attention}: 

(1) \textit{The Stanford Sentiment Treebank (SST)} \cite{socher2013recursive}, a collection of sentences tagged with sentiment on a discrete scale from 1 to 5, where 1 is the most negative and 5 the most positive. We omit the neutral class (3) and conflate scores 1 and 2 as well as 4 and 5 into negative and positive class, respectively; 

(2) \textit{IMDB Large Movie Reviews Corpus} (IMDB) \cite{maas-etal-2011-learning}, a binary sentiment classification dataset of movie reviews; 

(3) \textit{AG News Corpus}, a categorized set of news articles from various sources. We limit ourselves to binary classification between articles labelled as \textit{world} (0) and \textit{business} (1); 

(4) \textit{20 Newsgroups} similarly, we consider the task of discriminating between \textit{baseball} (0) and \textit{hockey} (1) in this dataset of newsgroup correspondences labelled with 20 categories; 

(5,6) \textit{MIMIC ICD9} \cite{johnson2016mimic}, a dataset of patient discharge summaries from a database of electronic health records. Here, we analyse two classification tasks on different subsets of the data: whether a summary is labelled with the ICD9 code for \textit{diabetes} (1) or \textit{not} (0) (henceforth \textit{Diabetes}) and whether a summary corresponds to a patient with \textit{acute} (0) or \textit{chronic} anemie (henceforth \textit{Anemia}); 

For consistency, we use the train/test/dev splits produced by \newcite{jain2019attention}.\footnote{\url{https://github.com/successar/AttentionExplanation}}

\section{Results} %
\label{sec:experiments}

\subsection{Attention is Fragile} %
\label{sub:attention_is_fragile}

We report the average F1-scores of five runs for the \textbf{base} model and the following regularization variants: \textbf{concat}, \textbf{tying}, and \textbf{MLM}.
We omit results on \textbf{residual} due to space, but they are consistently comparable to \textbf{concat} due to their similar nature.
For each model variant we report results of experiments with the dot-product ($\bigcdot$) and additive (+) attention mechanism.
Due to space constraints, we omit the full results and refer the reader to Appendix for more details.

We report the performance of each model in scenarios where we use trained attention (\textbf{Tr.}), a random permutation of the trained attention (\textbf{Pm.}) or substitute the attention distribution with the uniform (\textbf{Un.}).
For the uniform and permutation settings, we report the drop in F1-score when compared to trained attention performance.

\begin{table*}[t]
{\small
\centering
\begin{tabular}{|c|c|c|c|c|c|c|c|c|c|c|c|c|c|}
\cline{3-14}
\multicolumn{2}{c}{} & \multicolumn{3}{|c|}{Base} & \multicolumn{3}{|c|}{Concat} & \multicolumn{3}{|c|}{Tying} & \multicolumn{3}{|c|}{MLM} \\
\cline{2-14}
\multicolumn{1}{c|}{} & $\alpha$ & $\uparrow$ Tr. & $\downarrow$ Un. & $\downarrow$ Pm. & $\uparrow$ Tr. & $\downarrow$ Un. & $\downarrow$ Pm. & $\uparrow$ Tr. & $\downarrow$ Un. & $\downarrow$ Pm. & $\uparrow$ Tr. & $\downarrow$ Un. & $\downarrow$ Pm. \\
\hline
\multirow{ 2}{*}{SST} & + & $\mathbf{84.2}$ & $-2.8$ & $-5.0$  &$83.7$ & $-1.9$ & $-4.7$ & $83.5$ & $\mathbf{-7.0}$ & $\mathbf{-18.0}$ & $82.8$ & $-5.9$ & $-15.6$ \\
 & $\bigcdot$ & $\mathbf{84.3}$ & $-2.6$ & $-6.5$& $84.1$ & $-3.4$ & $-7.4$ & $83.5$ & $\mathbf{-9.9}$ & $\mathbf{-20.0}$  & $82.7$ & $-3.0$ & $-5.4$ \\ \hline

\multirow{ 2}{*}{AG} & + & $\mathbf{95.9}$ & $-2.3$ & $-3.9$ &$\mathbf{95.9}$ & $-1.5$ & $-2.6$ & $95.0$ & $\mathbf{-3.3}$ & $\mathbf{-14.6}$  & $95.2$ & $-1.5$ & $-6.6$ \\
 & $\bigcdot$ & $95.9$ & $-2.3$ & $-3.8$  & $\mathbf{96.1}$ & $-1.9$ & $-3.0$  & $95.4$ & $\mathbf{-3.0}$ & $\mathbf{-12.2}$  & $95.4$ & $-2.0$ & $-5.3$ \\ \hline

\multirow{ 2}{*}{NG} & + & $90.9$ & $-9.8$ & $-14.1$ & $91.3$ & $-25.0$ & $-28.2$  & $91.4$ & $-39.7$ & $-43.2$  & $\mathbf{91.5}$ & $\mathbf{-76.3}$ & $\mathbf{-66.0}$\\
 & $\bigcdot$ & $\mathbf{91.1}$ & $-35.2$ & $-36.8$ & $91.0$ & $-40.4$ & $-37.1$  & $90.9$ & $-37.0$ & $-42.8$ & $89.1$ & $\mathbf{-79.6}$ & $\mathbf{-72.8}$ \\ \hline

\multirow{ 2}{*}{IM} & + & $\mathbf{88.3}$ & $-10.0$ & $-13.4$  & $\mathbf{88.3}$ & $-10.2$ & $-14.0$  & $87.1$ & $\mathbf{-56.2}$ & $\mathbf{-43.3}$  & $87.5$ & $-22.8$ & $-26.5$ \\
 & $\bigcdot$ & $\mathbf{88.2}$ & $-18.6$ & $-22.9$ & $87.9$ & $-17.2$ & $-20.8$  & $87.2$ & $\mathbf{-57.7}$ & $\mathbf{-45.3}$  & $87.8$ & $-15.3$ & $-18.5$ \\ \hline

\multirow{ 2}{*}{ANM} & + & $92.4$ & $-21.6$ & $-22.4$  & $\mathbf{92.8}$ & $-19.3$ & $-22.2$  & $91.3$ & $-31.4$ & $-27.6$  & $89.7$ & $\mathbf{-35.0}$ & $\mathbf{-37.7}$ \\
 & $\bigcdot$ & $\mathbf{92.7}$ & $-10.2$ & $-14.4$  & $92.4$ & $-15.2$ & $-17.2$  & $91.0$ & $\mathbf{-91.0}$ & $\mathbf{-59.7}$  & $90.7$ & $-37.8$ & $-33.9$ \\ \hline

\end{tabular}
}
\caption{\%\,F1-scores for trained models (higher is better) and drops in performance ($\Delta$ F1) when applying regularization (lower is better). Scores reported are averages over five runs.}
\label{tab:short_results}
\end{table*}

We omit the results on the Diabetes dataset, as every modification of attention weights on this dataset results with an F1-score of $0$, due to a very small number of tokens being a high-precision indicator of the positive class, as noted by \newcite{jain2019attention}.
As shown in Table~\ref{tab:short_results}, regularization setups increase fragility of model performance with respect to modifications of the attention distribution, while retaining similar classification scores to the base model.
These results indicate that we have successfully reduced the space of possible alternative explanations for the model by tying the input and hidden representations closer together.
By doing this, we show that lateral information leakage (between hidden states) is reduced when proper regularization is applied, and that, as a consequence, alternative explanations are also plausible.
Having shown this, we still need to determine whether a high attention weight on a hidden state is a faithful measure of importance of a corresponding input.

\subsection{Mutual Information is Higher} %
\label{sub:mutual_information_is_higher}

In Table~\ref{tab:mi_results} we report mutual information scores across datasets for the most frequent words (MF) and a random sample drawn from the scaled unigram distribution of the vocabulary (RS).

The increase in mutual information scores between inputs $x_t$ and hidden states $h_t$ implies that more information from the inputs is retained during encoding.
While retention of input information is not a desirable trait of a model performing pure sequence classification, as the only goal the model optimizes is producing the correct class label with high confidence, it is beneficial for interpretability.
If we wish to interpret classifier decisions through inspecting attention weights on hidden states, we have to ensure that a hidden state preserves a significant degree of information from the input.
A significant increase in mutual information suggests that the base model was filtering or overwriting a large amount of information from the input, making attention inspection less credible.
It is not possible to report mutual information for the \textbf{concat} setup as the dimensionality of the hidden vector is larger than the input embedding, so we report the results for \textbf{Residual}. The results for the Residual setup can be considered close to the best realistically obtainable MI score as the model explicitly includes the input embedding in the hidden state.

\begin{table}[t]
\centering
\begin{tabular}{|c|c|c|c|c|c|}
\cline{2-6}
\multicolumn{1}{c|}{} & $\sim$ & Base & Resid & Tying & MLM  \\
\hline

\multirow{ 2}{*}{SST} & MF  & $2.324$ & $\mathbf{5.062}$ & $4.870$  & $3.662$ \\
& RS & $2.435$ & $\mathbf{4.289}$ & $4.216$  & $3.808$ \\ \hline
\hline

\multirow{ 2}{*}{AG} & MF & $1.940$ & $\mathbf{5.467}$ & $4.075$ & $3.845$ \\
& RS& $2.078$ & $\mathbf{4.518}$ & $4.177$ & $3.980$  \\  \hline
\hline

\multirow{ 2}{*}{NG} & MF  & $1.566$ & $\mathbf{4.345}$ & $3.985$ & $3.677$ \\
& RS  & $1.828$ &  $\mathbf{3.843}$ & $3.784$ & $3.458$ \\  \hline
\hline

\multirow{ 2}{*}{IM} & MF  & $2.455$ & $4.998$ & $\mathbf{5.186}$ & $3.728$ \\
& RS & $2.682$ & $\mathbf{4.366}$ & $4.434$ & $3.885$ \\ \hline
\hline

\multirow{ 2}{*}{ANM} & MF & $3.711$ & $\mathbf{5.253}$ & $4.239$  & $4.016$ \\
& RS  & $3.780$ & $\mathbf{4.477}$ & $3.950$ & $3.921$ \\  \hline%

\end{tabular}
\caption{Mutual information scores between the input and hidden representations. Higher is better. Due to space limitations, results are only reported on additive attention.}
\label{tab:mi_results}
\end{table}

\subsection{Adversarial Attention Distributions are Harder to Find} %
\label{sub:it_is_harder_to_find_counterfactual_attention_distributions}

\begin{figure}
\centering
  \includegraphics[width=0.42\textwidth]{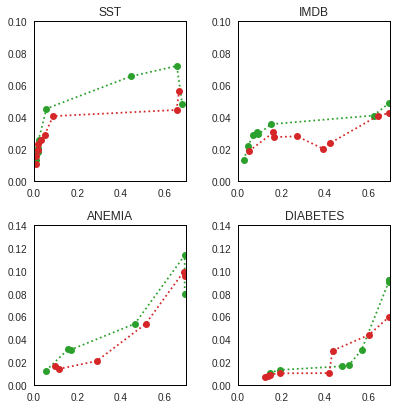}
  \caption{Averaged per-instance test set JSD (x-axis) and TVD (y-axis)}
  \label{fig:tvd_jsd}
\end{figure}

In Fig.~\ref{fig:tvd_jsd} we report results where for a fixed oracle model we train an adversary with the objective of minimizing the TVD between the predictions of the model and, at the same time, maximizing JSD between per-instance averaged attention distributions.
Due to space limitations, we only report results for the MLM regularised model, while the others fare comparably.
The red dotted line indicates the imitation setup of the base model, and the green dotted line indicates imitation setup for the MLM model.
Consistently, except for an outlier point in the Diabetes dataset, the imitation setup of the MLM model produces larger drops of TVD in order to increase the JSD between attention distributions, corroborating the claim that attention distribution of the MLM model is more fragile.

\begin{figure*}
\centering
  \includegraphics[width=\textwidth]{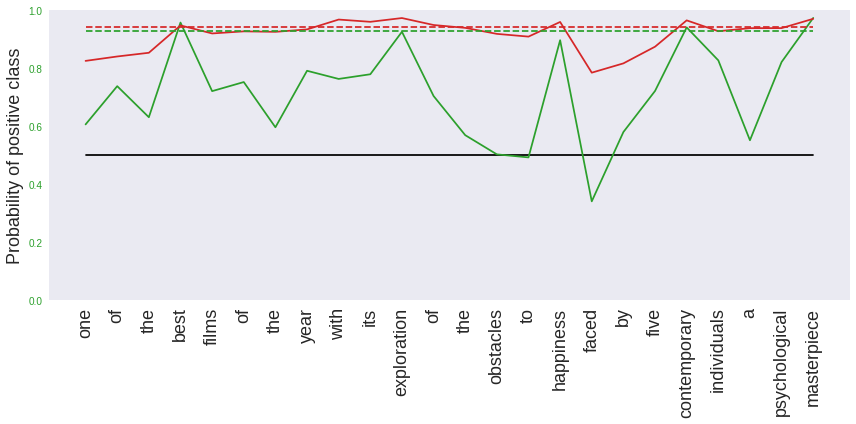}
  \caption{Per-token prediction probability for an example from the SST dataset for the base model (red) and a regularized (tying) model (green).
  The dotted lines indicate the classification probability of the model.
  More instances and examples of other regularization techniques can be found in the Appendix.}
  \label{fig:hyperplane_dist}
\end{figure*}

\section{Understanding the Effect of Model Regularization} %
\label{sec:effect_of_masked_language_modelling}

To visually demonstrate the undesired effect of attention mechanisms when trained in the \textbf{base} setting, as well as to illustrate the effect of regularizations we applied, we first analyse how we obtain the classifier prediction.
The output of the classifier is an affine transformation of the attention output:
\begin{equation}
\begin{split}
p_{\mathit{logit}} &= W_d (\sum_{i=1}^T \alpha_i h_i) + b_d \\
            &= \sum_{i=1}^T \alpha_i (W_d h_i + b_d)  \\
            &= \sum_{i=1}^T \alpha_i \hat{p}_t
\end{split}
\end{equation}
We can reformulate this as a convex attention-weighted sum of logits ($\hat{p}_t$) obtained by running each individual hidden state through the decoder.
Once we scale the logits for individual timesteps, we obtain the prediction probability as if the whole attention mass was on that hidden representation.
For attention weights to be a faithful measure of interpretability, this probability should be high only on tokens which are decision-relevant.

In Fig.~\ref{fig:hyperplane_dist}, we plot these token-level probabilities for a single example to demonstrate that in the \textbf{base} model, this is not the case.
We can see that for the base model, the probabilities for most tokens have nearly the \textbf{same} probability as the final prediction, while the regularization keeps the representations for neutral words grounded closer to the decision boundary.
As a direct result of this, the model predictions are much more fragile to change of attention weights, as only a small number of hidden states are far enough from the decision boundary to produce an equally confident classification.

We now quantitatively formulate and measure this criterion -- \textbf{if} the accuracy of a regularized classifier isn't hurt by the regularization, when optimizing for interpretability we should prefer models that have a lower per-token average prediction probability (given that the prediction for that instance is correct).

\begin{table}[t]
\centering
\begin{tabular}{|c|c|c|c|c|c|}
\cline{3-6}
\multicolumn{2}{c|}{} & Base & Resid & Tying & MLM  \\
\hline

\multirow{ 2}{*}{SST} & + & $0.712$ & $0.685$ & $\mathbf{0.586}$ & $0.630$ \\ 
 & $\bigcdot$ & $0.693$ & $0.701$ & $\mathbf{0.600}$ & $0.664$ \\
 \hline

\multirow{ 2}{*}{AG} & + & $0.887$ & $0.822$ & $\mathbf{0.615}$ & $0.695$ \\ 
 & $\bigcdot$ & $0.862$ & $0.876$ & $\mathbf{0.646}$ & $0.698$ \\
 \hline

\multirow{ 2}{*}{NG} & + & $0.811$ & $0.551$ & $0.577$ & $\mathbf{0.514}$ \\
 & $\bigcdot$ & $0.687$ & $0.755$ & $0.516$  & $\mathbf{0.482}$ \\
\hline

\multirow{ 2}{*}{IM} & + & $0.625$ & $0.609$ & $\mathbf{0.533}$ & $0.562$ \\
 & $\bigcdot$ & $0.590$ & $0.608$ & $\mathbf{0.539}$ & $0.553$ \\
\hline

\multirow{ 2}{*}{AN} & +  & $0.568$ & $0.547$ & $0.531$ & $\mathbf{0.515}$ \\
 & $\bigcdot$ & $0.542$ & $0.534$  & $0.515$ & $\mathbf{0.519}$ \\
\hline

\end{tabular}
\caption{Average per-token prediction probability across models and tasks. From the perspective of interpretability, lower is better, given the classifier performance is not significantly affected.}
\label{tab:mi_results}
\end{table}

\section{Conclusion} %
\label{sec:conclusion}

We have identified the lack of a word-level objective as the likely cause of attention weights not being a faithful tool of interpretability in the case of sequence classification with attention mechanism augmented recurrent networks.
We experimentally establish that we can add regularization methods to sequence classification which strengthen the relationship between the input and hidden states while not being a detriment to classification performance.
If one wishes to interpret classifier decisions through inspecting attention weights, we strongly suggest inclusion of a technique such as weight tying or adding masked language modelling as an auxiliary.
Adding such methods causes the model to become more susceptible to attacks modifying the attention weights of a trained model, and increases faithfulness of explanations produced by attention weights.

While we believe our work is a step forward towards using attention weights as a faithful explanation, by no means do we claim that the modification is sufficient.
As was our primary concern, the risk with using attention weights as a tool of interpretability is that a single bad explanation could have consequences in decision-making scenarios, and while our methods improve the faithfulness of such interpretability, it is by no means foolproof.
We have only scratched the surface of faithful interpretability, and most of the datasets in our and previous work do not have human annotated rationales.
In order to fully understand the cases in which attention provides a reliable explanation, we believe that datasets with annotated rationales or decision-relevant tokens should be used.
This analysis should also be extended to more complex models which better capture the nuances of language. 
We believe that the experiments we presented demonstrate the shortcomings of interpreting model decisions through inspecting attention weights, however  we acknowledge that this branch of research sorely lacks evaluation methods that include humans in the loop.

\bibliography{anthology,acl2020}
\bibliographystyle{acl_natbib}

\clearpage
\appendix
\section{Model Hyperparameters}

Since we analyse a number of models and regularization techniques, we naturally also have a large number of hyperparameters.
We do not tune any of them except for regularization-specific ones and we inherit others them from previous work \cite{jain2019attention, wiegreffe2019attention}.
A notable change is the dimension of the hidden state, which we increase from 128 to 150 due to the nature of the MLM regularization.
We, however, repeat the experiments for the base model with this increased dimensionality.

\begin{table}[t]
\centering
\begin{tabular}{|l|r|}
\hline
\multicolumn{2}{|c|}{General parameters}\\
\hline
Embedding dim & $300$ \\
RNN hidden dim & $150$ \\
Learning rate & $1\mathrm{e}{-3}$ \\
Grad. clipping & $5$ \\
Batch size & $32$ \\
Weight decay & $1\mathrm{e}{-5}$ \\
\hline
\multicolumn{2}{|c|}{Regularization parameters}\\
\hline
Masking prob. & $0.15$ \\
Masking weight $\eta$ & $\{0.1,0.3,1,3,5\}$ \\
Tying weight $\delta$ & $\{10, 20, 30\}$ \\
\hline
\end{tabular}
\caption{Model hyperparameters}
\label{tab:params}
\end{table}

We report our parameters in Table~\ref{tab:params}.
While we have considered other values in a brief search for $\eta$ and $\delta$, but we have only ablated over the mentioned ones as they have proven to be (locally) optimal.

\begin{table}[h]
\centering
\begin{tabular}{|l|r|r|}
\hline
Dataset & Avg. len. & Vocabulary \\ \hline
SST & 17 & 17310 \\
AG News & 31 & 15286 \\
20NG & 164 & 15590 \\
IMDB & 234 & 41919 \\
Diabetes & 1700 & 23778 \\
Anemia & 1927 & 20290 \\
\hline
\end{tabular}
\caption{Statistics of datasets used in experiments}
\label{tab:stats}
\end{table}

We also report the statistics of datasets used in experiments in Table~\ref{tab:stats}. The average instance length had a significant impact on the experiments as datasets with longer instances were naturally more fragile to attention distribution modifications.

\section{Experiments on Multilayer LSTMs} %
\label{sec:experiments_on_multilayer_lstms}

All of the experiments performed in the paper used single-layer LSTMs.
Even though the considered binary classification tasks could be considered some of the simplest NLP problems, one still wonders what would the effect be if a more complex encoder was used.
To this end, we perform a preliminary set of experiments where we use the best hyperparameters used for training of the single-layer networks and increase the number of layers of the LSTM network.

\begin{table*}[t]
{\small
\centering
\begin{tabular}{|c|c|c|c|c|c|c|c|c|c|c|c|c|c|}
\cline{3-14}
\multicolumn{2}{c}{} & \multicolumn{3}{|c|}{Base} & \multicolumn{3}{|c|}{Concat} & \multicolumn{3}{|c|}{Tying} & \multicolumn{3}{|c|}{MLM} \\
\cline{2-14}
\multicolumn{1}{c|}{} & \#L & $\uparrow$ Tr. & $\downarrow$ Un. & $\downarrow$ Pm. & $\uparrow$ Tr. & $\downarrow$ Un. & $\downarrow$ Pm. & $\uparrow$ Tr. & $\downarrow$ Un. & $\downarrow$ Pm. & $\uparrow$ Tr. & $\downarrow$ Un. & $\downarrow$ Pm. \\
\hline
\multirow{ 3}{*}{SST} & $1$ & $\mathbf{84.2}$ & $-2.8$ & $-5.0$  &$83.7$ & $-1.9$ & $-4.7$ & $83.5$ & $\mathbf{-7.0}$ & $\mathbf{-18.0}$ & $82.8$ & $-5.9$ & $-15.6$ \\
 & $2$ & $84.2$ & $-1.0$ & $-1.2$  & $\mathbf{84.5}$ & $-0.8$ & $-4.6$ & $84.1$ & $-3.7$ & $\mathbf{-14.5}$ & $84.4$ & $\mathbf{-3.8}$ & $-5.9$ \\
 & $3$ & $84.2$ & $-0.7$ & $-0.7$ & $83.3$ & $-1.3$ & $-1.3$ & $\mathbf{84.6}$ & $-2.7$ & $-13.9$  & $82.3$ & $\mathbf{-7.3}$ & $\mathbf{-18.0}$ \\ \hline

\multirow{ 3}{*}{AG} & 1 & $\mathbf{95.9}$ & $-2.3$ & $-3.9$ &$\mathbf{95.9}$ & $-1.5$ & $-2.6$ & $95.0$ & $\mathbf{-3.3}$ & $\mathbf{-14.6}$  & $95.2$ & $-1.5$ & $-6.6$ \\
 & $2$ & $95.7$ & $-0.3$ & $-0.3$ &$\mathbf{95.9}$ & $-1.4$ & $-2.0$ & $95.5$ & $\mathbf{-3.7}$ & $\mathbf{-14.8}$  & $95.6$ & $-1.6$ & $-3.8$ \\
 & $3$ & $\mathbf{95.9}$ & $-0.0$ & $-0.1$  & $95.7$ & $-1.0$ & $-1.6$  & $95.4$ & $-2.0$ & $-12.8$  & $95.8$ & $\mathbf{-13.2}$ & $\mathbf{-62.5}$\\ \hline

\multirow{ 3}{*}{NG} & 1 & $90.9$ & $-9.8$ & $-14.1$ & $91.3$ & $-25.0$ & $-28.2$  & $91.4$ & $-39.7$ & $-43.2$  & $\mathbf{91.5}$ & $\mathbf{-76.3}$ & $\mathbf{-66.0}$\\
 & $2$ & $93.7$ & $-0.9$ & $-5.6$ & $\mathbf{94.0}$ & $-6.3$ & $-11.9$  & $92.8$ & $-17.5$ & $-25.5$ & $89.8$ & $\mathbf{-31.6}$ & $\mathbf{-35.0}$\\
 & $3$ & $\mathbf{92.0}$ & $0.0$ & $0.0$ & $91.5$ & $-30.2$ & $-35.7$  & $89.0$ & $\mathbf{-30.3}$ & $\mathbf{-39.3}$ & $88.5$ & $-17.9$ & $-17.4$ \\ \hline

\multirow{ 3}{*}{IM} & 1 & $\mathbf{88.3}$ & $-10.0$ & $-13.4$  & $\mathbf{88.3}$ & $-10.2$ & $-14.0$  & $87.1$ & $\mathbf{-56.2}$ & $\mathbf{-43.3}$  & $87.5$ & $-22.8$ & $-26.5$ \\
 & $2$ & $88.4$ & $-3.1$ & $-3.8$ & $\mathbf{88.9}$ & $-7.2$ & $-9.1$  & $87.6$ & $\mathbf{-51.2}$ & $\mathbf{-41.1}$  & $87.4$ & $-14.5$ & $-21.7$ \\
 & $3$ & $88.5$ & $-1.2$ & $-1.4$ & $\mathbf{88.8}$ & $-5.7$ & $-7.8$ & $87.9$ & $-7.6$ & $-21.3$  & $87.1$ & $\mathbf{-87.1}$ & $\mathbf{-84.5}$\\ \hline

\end{tabular}
}
\caption{\%\,F1-scores for trained models (higher is better) and drops in performance ($\Delta$ F1) for LSTM models with multiple layers. The number of layers is indicated in the second column.}
\label{tab:layer_results}
\end{table*}

The results in Table~\ref{tab:layer_results}, while far from conclusive, show that (1) among all tasks, the base model consistently becomes \textbf{more robust} to attention perturbation the more layers we add.
Inconsistently, we further observe a (2) \textbf{diminishing return} of regularization techniques among tasks as the number of layers increases. In some cases, the 3-layer results do not follow this trend (but, curiously, the regularization seems to have a stronger effect). 
We believe that these results should be taken with a grain of salt prior to a careful ablation study, but still might interest the reader.

\section{Importance of Initialisation in Dot-Product Attention}

Initially, the experiments we conducted worked well for additive attention but not for scaled dot-product attention.
While the various regularization techniques produced significant changes in F1-scores when the additive attention distribution was modified post-hoc, this was not the case for dot-product attention and the F1-scores remained constant no matter the modification.
This was caused by the fact that the attention distribution of the model consistently converged to a uniform one.

After exhaustive experimenting, the only change that fixed this behavior was changing the default initialization scheme for the query parameter.
The dot-product self-attention mechanism for a \textbf{single instance} (for illustrative purposes) is generally defined as follows:
\begin{equation}
\mathrm{Attention}(q,K,V) = \mathrm{softmax}(\frac{q K^T}{\sqrt{d_k}}) V
  \label{eq:dot_attn}
\end{equation}
where $q$ is the query vector, while $K$ and $V$ are stacked representations for each timestep.
In practice, when using self-attention for single-sequence classification, the query is a model parameter,\footnote{This independence of the query vector from the instance is not intuitive in our perspective (it seems natural to us that different information is relevant for different instances), but in practice we find that both approaches work equally well.} while the keys and values are functions of RNN hidden states.
In our case concretely (following \newcite{jain2019attention, wiegreffe2019attention}), the keys and values are the hidden states themselves.

With this in mind, Eq.~\ref{eq:dot_attn} can be written as follows:
\begin{equation}
\mathrm{Attention}(H) = \mathrm{softmax}(\frac{L_q(H)}{\sqrt{d_k}}) H
\end{equation}
where $L_q$ is the trainable query parameter.
In our \verb|Pytorch| implementation, $L_q$ is a \verb|Linear| layer, which is initialised from the Kaiming uniform\footnote{\url{https://github.com/pytorch/pytorch/blob/master/torch/nn/modules/linear.py\#L79}} distribution with the scale parameter $\sqrt{5}$.
With this initialisation, the dot-product attention distribution in our experiments has always converged to a uniform one.
When we changed the initialisation to instead sample from a standard normal distribution, the dot-product attention converges to a sensible distribution.
We suspect this problem occurs because the small initial weights of the linear transform scale down the difference norm between the attention probabilities too much to be distinguished from the uniform distribution.

\section{Additional Visualisations of Regularization Effects}

\begin{figure}
\centering
  \includegraphics[width=0.5\textwidth]{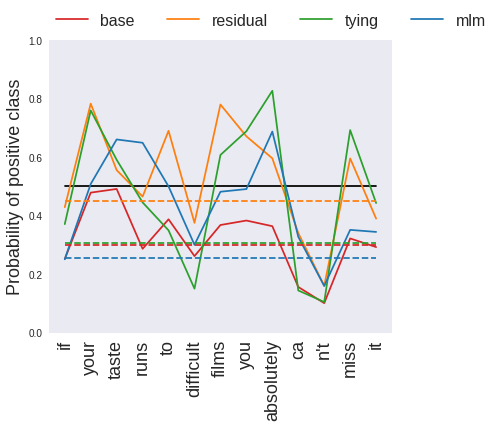}
  \caption{A negative example: perhaps the analysed single-layer LSTM is unable to understand even the simple nuances of language. Here the instance is classified as negative across all models only due to presence of the word ``difficult''. Note that these models obtain a near 0.9 F1-score on this dataset.}
  \label{fig:example_1}
\end{figure}

\begin{figure}
\centering
  \includegraphics[width=0.4\textwidth]{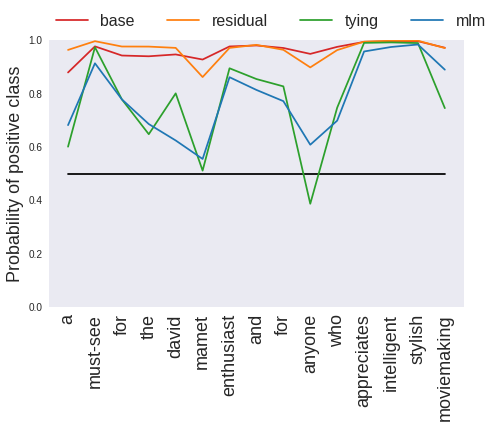}
  \caption{A clear-cut instance }
  \label{fig:example_2}
\end{figure}

\begin{figure*}
\centering
  \includegraphics[width=\textwidth]{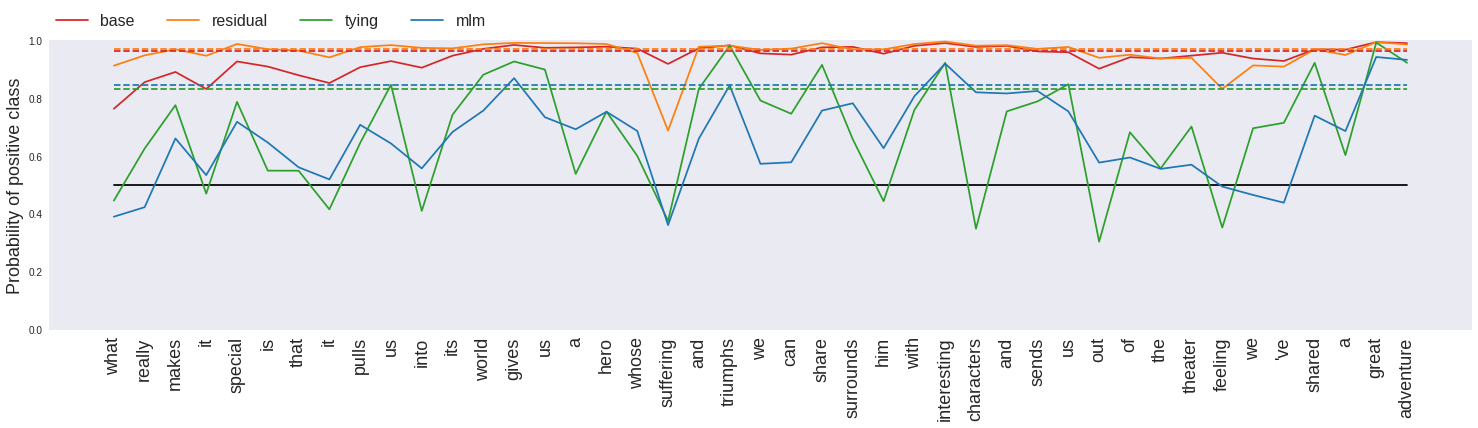}
  \caption{A long example which further demonstrates lateral information leakage}
  \label{fig:example_3}
\end{figure*}

\begin{figure}
\centering
  \includegraphics[width=0.5\textwidth]{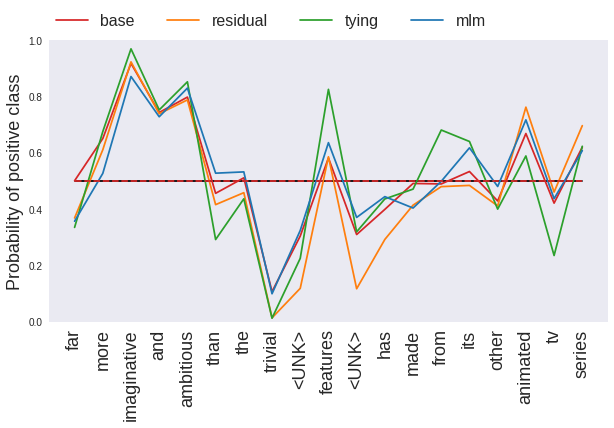}
  \caption{We observe that for instances where the model is not clear about the classification, the per-word probabilities are pretty similar between regularizations. We believe that lateral information leakage happens only when the model is confident in its prediction. Base model prediction confidence is indicated in this example (it overlaps with the $0.5$ line).}
  \label{fig:example_4}
\end{figure}

\begin{figure}
\centering
  \includegraphics[width=0.3\textwidth]{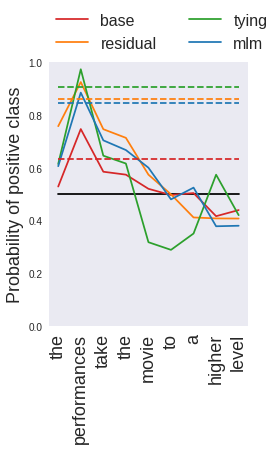}
  \caption{A rare example where the regularised models are more confident in the correct prediction than the base model}
  \label{fig:example_5}
\end{figure}

To expand on Fig.~\ref{fig:hyperplane_dist}, we now plot per-token prediction probabilities for multiple models. We sometimes omit the model classification probabilities not to clutter the plots too much.
We select diverse examples (Figs.~\ref{fig:example_1}--\ref{fig:example_5}) from the first three batches of the SST validation split.

\end{document}